\pdfoutput=1

\documentclass[11pt]{article}

\usepackage[final]{acl}

\usepackage{times}
\usepackage{latexsym}

\usepackage[T1]{fontenc}

\usepackage[utf8]{inputenc}

\usepackage{microtype}

\usepackage{inconsolata}

\usepackage{graphicx}

\usepackage{amsmath}
\usepackage{amsthm}
\usepackage{booktabs}
\usepackage{algorithm}
\usepackage{algorithmic}

\title{CA-SQL: Complexity-Aware Inference Time Reasoning for Text-to-SQL via Exploration and Compute Budget Allocation}

\author{James Petullo \\
  Brandeis University / Waltham, MA\\
  \texttt{jamespetullo@brandeis.edu} \\\And
  Nianwen Xue\\
  Brandeis University / Waltham, MA\\
  \texttt{xuen@brandeis.edu} \\}

\begin{document}
\maketitle
\begin{abstract}
While recent advancements in inference-time learning have improved LLM reasoning on Text-to-SQL tasks, current solutions still struggle to perform well on the most challenging tasks in the Bird-Bench (BIRD) benchmark. This is due to inadequate solution space exploration, which is necessary to uncover promising candidate queries that can be further refined to produce the correct output. To address this challenge, we introduce CA-SQL, a novel Text-to-SQL pipeline that utilizes the estimated difficulty of a task to dynamically scale the breadth of the exploration for generating solution candidates. In addition, we use a custom prompt seeding method, based on principles of evolutionary search, to further elicit exploratory behavior from the base LLM and a novel voting method to select the best candidate solution at the end of the search. Experiments demonstrate that our solution achieves a state-of-the-art score of \textbf{51.72\%} on the “challenging” tier of BIRD development set problems, using only GPT-4o-mini, out-performing other in-context learning approaches, even those that leverage larger models. Overall, our method attains a competitive \textbf{61.06\%} execution accuracy and \textbf{68.77\%} Soft F1 score on the BIRD development dataset.
\end{abstract}

\section{Introduction}

\begin{figure}
\centering
\includegraphics[width=0.5\textwidth]{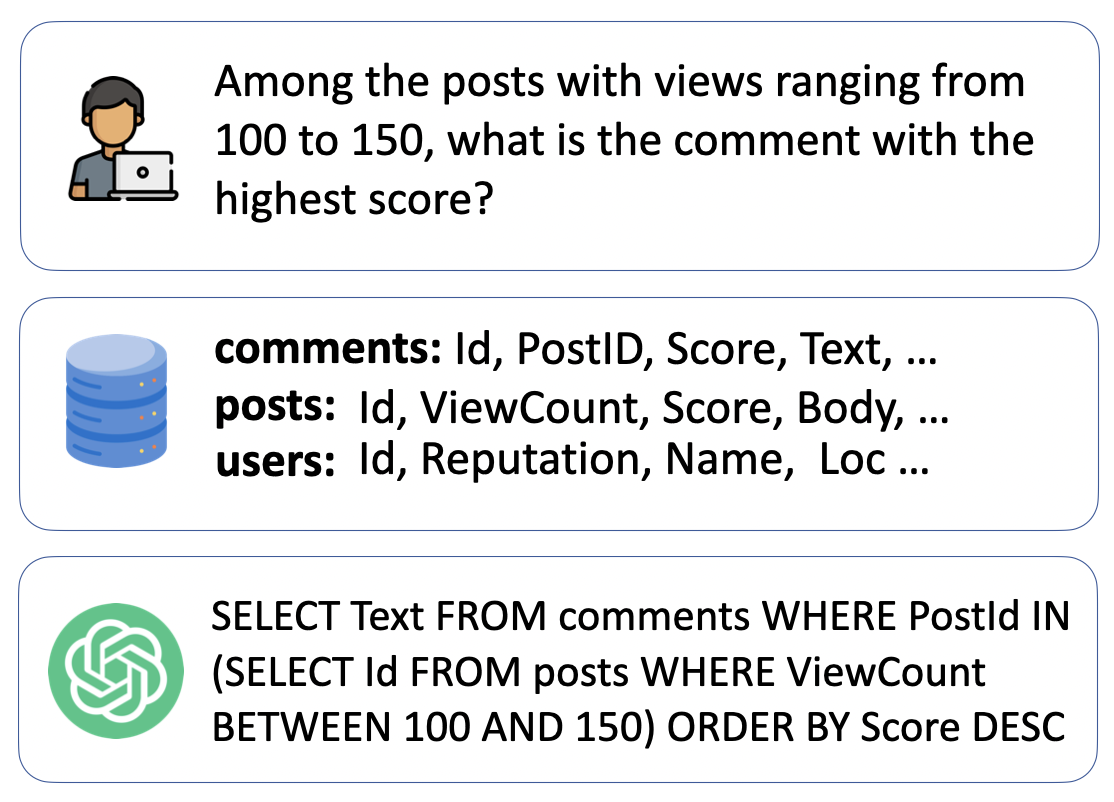}
\caption{The Text-to-SQL process, whereby a user's question, in conjunction with the full database schema, must be utilized by the LLM to produce an output query which answers the question.}
\label{fig:texttosql}
\end{figure}

A Text-to-SQL pipeline converts a user’s query task, in natural language format, to a valid SQL query that can return the user’s desired output when run on a database. The Text-to-SQL problem is uniquely difficult, for not only must an input task be decomposed and reformed as a structured query, but the schematics of the database, including its tables, columns, value types, and more must be accounted for during the query generation process to produce valid, accurate results (see Figure \ref{fig:texttosql}). While long the focus of semantic parsing techniques \cite{Eyal2023-et,Wang2022-ik,vougiouklis-etal-2023-fastrat}, the rise of LLMs have accelerated the development of Text-to-SQL algorithms that can rival and even exceed gold reference queries. LLMs, however, still struggle on more challenging Text-to-SQL tasks, characterized by greater ambiguity in output requirements and requiring a higher degree of candidate query complexity to properly solve the task. Current state-of-the-art in-context learning solutions exhibit impressive accuracy on problems of simple and moderate difficulty, but lag behind in performance on challenging tasks. The primary reason for a lack of commensurate performance on more complex tasks is due to a lack of adequate exploration in the candidate query search process: most Text-to-SQL solutions attempt to maintain a tradeoff between exploration of candidate queries and exploitation of the most promising candidates, however, our experiments demonstrate that popular query sampling methods for Text-to-SQL exploration yield a pool of candidates that possess a suboptimal number of unique queries. This, in turn, makes it less likely that the correct solution will be found and more costly to compensate for, as increasing the number of unique candidates in a solution pool necessitates additional, repeated calls to the query generation LLM.

Eliciting exploratory behavior in LLMs is a well-documented challenge \cite{Pan2025-kw,Huang2024-eq,Krishnamurthy2024-pp,LIGHT}. LLMs do not sufficiently explore their environment without frequent changes to the composition of their prompts \cite{LIGHT}, and proposed techniques to generalize exploration across problem domains, such as temperature increases or character role playing system prompts, frequently fail to provide adequate coverage of the search space \cite{Krishnamurthy2024-pp}.

To overcome exploration deficiencies in the Text-to-SQL realm, we propose a novel method of scaling the inference-time search breadth via a task difficulty score and devise a more efficient and effective prompt seeding technique to sample from the query space. To the best of our knowledge, we are the first to introduce a difficulty metric to guide the task solution generation process and empirically analyze existing Text-to-SQL exploration techniques and expound upon their limitations.

The standard Text-to-SQL pipeline consists of the following steps: (1) schema linking, whereby the most relevant tables and columns from the database are chosen for inclusion in the query generation prompt, (2) query generation from the selected schema, (3) refinement of the candidate queries to eliminate errors and correct any inaccuracies, and (4) the selection of a final solution query from the pool of candidates. Our work has identified several limitations pertaining to the first and fourth stages of the process:

\begin{itemize}
    \item Most solutions derive a population of candidate queries or expand a root solution (in the case of MCTS \cite{MCTS-MAIN}) from a single subset of columns and tables from the database schema. Even using high model temperatures and a random ordering of schema subset contents in the prompt, this approach leads to a loss of diversity in the candidate pool and limits the solution’s coverage of the query search space.
    \item These solutions do not dynamically scale to meet the complexity of the input task; that is, the same compute budget is allocated to each task, regardless of its difficulty. This means that candidate solutions for more challenging tasks may be inadequately explored or refined, while simpler problems may be overcorrected, leading to a degradation in output quality.
    \item In most settings and benchmarks, including BIRD, the gold reference query for a task is unknown or not made known to the Text-to-SQL pipeline. As such, most solutions rely on LLM-as-a-judge approaches or self-consistency voting methods to choose the best candidate. However, LLM-as-a-judge selectors have been shown to fail when attempting to choose between different queries, as they often cannot distinguish between subtitles in the queries’ composition and make more accurate selections. In addition, our experiments demonstrate that simple majority voting leads to poor performance on the BIRD development dataset.
\end{itemize}

\begin{figure}
\centering
\includegraphics[width=0.5\textwidth]{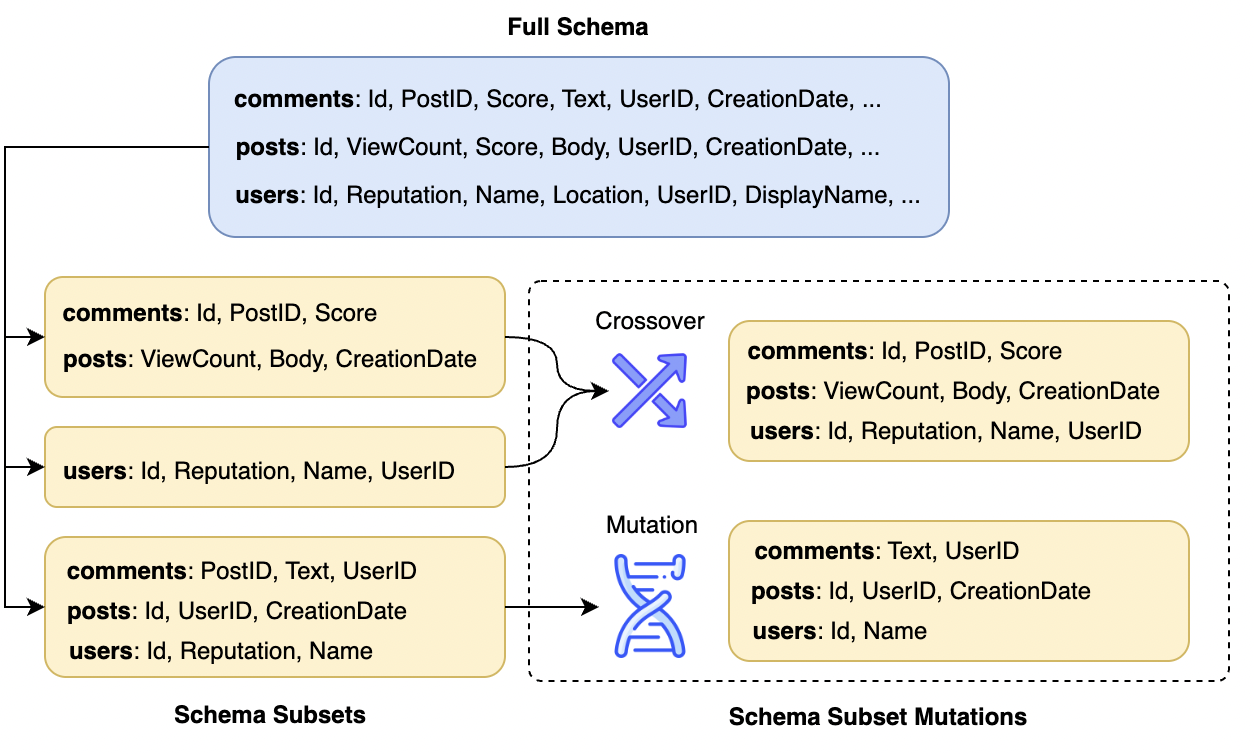}
\caption{Schema subsets are derived from the full database schema. Evolutionary operations of crossover and mutation are used to recombine schema subsets, producing novel table-column combinations which can be later used to seed candidate query prompts. Crossover is performed by merging the contents of two schema subsets, while mutation is a process of randomly eliminating table-column pairs.}
\label{fig:schemasubsets}
\end{figure}

In response to these limitations, we propose CA-SQL and make the following contributions:

\begin{itemize}
    \item  We devise a method of dynamically controlling the search breadth and depth via a complexity score, derived from an analysis of the task and its schema, to scale the search for candidate queries in a budget-aware fashion. 
    \item To expand the breadth of the candidate query search, we utilize multiple different schema subsets as seeds for the candidate queries, as opposed to one large subset from the same schema. Furthermore, we use the evolutionary search operators of crossover and mutation to expand this pool for subsequent exploration runs, eliminating the need to resample from the LLM subset generator. We demonstrate through experiments that this approach results in a greater uniqueness of candidate queries, as opposed to repeatedly sampling from an LLM provided with a single, larger subset. Please see figure \ref{fig:schemasubsets} for an example of the schema subset and mutation process.
    \item We propose a novel solution selection method, based on accumulated candidate evaluation scores, that outperforms standard self-consistency techniques, such as majority voting (see Section \ref{queryscore}). 
    \item We demonstrate that our solution exhibits increased performance on the “Challenging” task category of the BIRD benchmark \cite{Li2023-ca}, scoring a \textbf{51.72\%} execution accuracy when using GPT-4o-mini \cite{openai_gpt4omini_2024} as our primary LLM model. This outperforms all existing in-context learning methods on that problem category, including those relying upon larger, more powerful models, such as GPT-4 and GPT-4o. Overall, we score a competitive \textbf{61.06\%} execution accuracy and \textbf{68.77\%} Soft-F1 score on the BIRD development set. In this way, our approach closes the gap that previously existed between the most challenging problems and the rest.
\end{itemize}

\section{Related Work}

This section contains an overview of LLM-based advances in Text-to-SQL processing and its connections to the domain of LLM inference-time reasoning, from which our work takes its inspiration. We discuss current challenges in inference-time reasoning, especially in LLM agent environment exploration and discuss how our solution is designed to address those issues.

\subsection{Text-to-SQL}

The advent of large language models (LLM) has greatly improved the capabilities of current Text-to-SQL pipelines. Recent work has harnessed the advanced reasoning capabilities of LLMs to perform remarkably well on Text-to-SQL benchmarks, including the popular BIRD benchmark. These solutions can be categorized by the method they use to enhance their abilities on Text-to-SQL problems, either through training-time reasoning or inference-time reasoning. The first approach consists of fine-tuning a base model to improve its zero-shot and few-shot reasoning capabilities for specific use on Text-to-SQL tasks \cite{Pourreza2025-nj,Ma2025-xk,Chen2024-wu,He2025-ly}, while the second approach is to use in-context learning to improve the quality of a base LLM model’s reasoning to generate a query for a task \cite{MCS,MCTS,DAIL,CHASE,OPENSEARCH}. As LLM fine-tuning is an expensive and time-consuming process, necessitating a large volume of high-quality training samples, in-context learning presents a more cost-effective approach that, when properly deployed, can frequently match and even outperform larger LLMs or custom fine-tuned models on the same benchmark tasks.

Formally, the Text-to-SQL pipeline $F$ receives an input $X=(Q, S, H)$, where $Q$ is a natural language question, $S$ is a set containing database schema values, and $H$ is a short usage example. A database’s schema consist of tables $T = T_{1}, T_{2},...,T_{|T|}$ and associated columns $C = C_{1},C_{2},...,C_{|C|}$. Each column $C_{i}$ contains supplementary information, including their datatypes $D_{i}$ and example values $V_{i} = e_{1},e_{2},...,e_{|e|}$. The schema set S consists of tuples $(T_{i}, C_{i,j}, D_{i,j}, V_{i,j})$ for each table $T_{i}$ and an associated column $C_{i,j}$. The result of $F(X)$ is a query $Y$ which, when run on the database, produces the correct output that solves $Q$.

\subsection{Inference-Time Learning}

In-context methods fall under the category of inference-time learning. Here, additional compute time is allocated for the base model to reason longer during inference, as opposed to more extensive training or fine-tuning. Common methods for improving LLM reasoning during inference include Chain-of-Thought (CoT) for LLM answer refinement \cite{Wei2022-zs}, repeated sampling via an LLM from the solution space \cite{Mialon2023-zq}; and self-consistency methods for choosing a final solution from a set of candidates \cite{Wang2022-wm,White2023-fm}. As such, inference-time learning possesses two primary facets: (1) breadth, consisting of the total number of parallel reasoning chains, seeded by sampling from the solution space, and (2) depth, the length of each individual reasoning path. Correctly leveraging inference-time learning techniques necessitates scaling the breadth and depth of the search to meet the complexity of the task. This constitutes the exploration-exploitation tradeoff, whereby promising solutions must first be discovered and then refined to produce the correct answer. Recent work has utilized Best-of-N \cite{DEEPMIND}, Beam Search \cite{DEEPMIND,Feng2023-uk}, and Monte-Carlo Tree Search (MCTS) \cite{Xie2024-er}, among others, to balance the exploration-exploitation tradeoff in inference-time learning contexts.

\subsection{Inference-Time Learning and Text-to-SQL}

Prior Text-to-SQL solutions have leveraged CoT \cite{DAIL,ESQL,OPENSEARCH}, Beam Search \cite{BEAM}, and Monte-Carlo Tree Search \cite{MCTS} to navigate the solution space and refine candidate queries. Maximum reward \cite{MCTS}, self-consistency voting \cite{MCS}, and LLM-as-a-judge techniques \cite{CHASE} have been used to choose a final solution query from a pool of candidates. 

As we discuss in later sections, the exploration of the query solution space is vital to improving an inference-time learning approach’s ability to solve Text-to-SQL problems, especially the more challenging ones. Depth-based solutions \cite{OPENSEARCH}, suffer from a lack of initial exploration, which stunts their performance on difficult benchmark problems. Solutions that attempt to balance the exploration-exploitation tradeoff via MCTS \cite{MCTS} and candidate pool generation \cite{MCS} perform better, but provide inadequate variance in the prompt of the sampling LLM, reducing the number of diverse, unique queries in the candidate pool. In addition, these solutions do not scale the rate of exploration in a fashion commensurate to the complexity of the task. This means that simpler problems receive an extraneous allocation of compute resources and suffer from overoptimization, while more complex tasks have solution spaces that go underexplored. We further examine these shortcomings in Section \ref{limmcts}.

\section{Methodology}

\subsection{Overview}

\begin{figure*}
\centering
\includegraphics[width=1\textwidth]{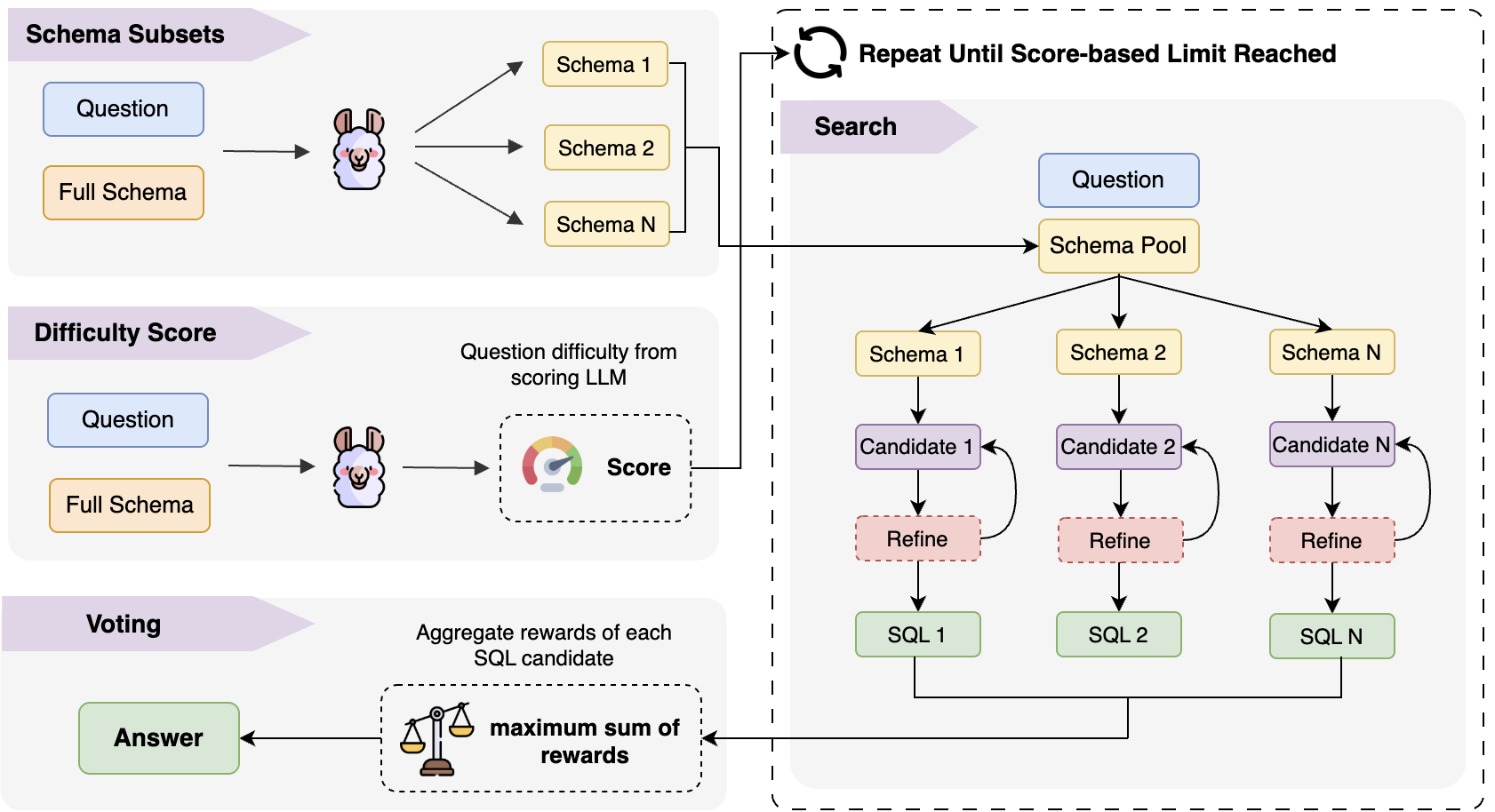}
\caption{Overview of CA-SQL's methodology. The core components of our proposed approach include schema subset generation, task difficulty scoring, search scaling, and final answer selection via sum-of-rewards voting.}
\label{fig:overview}
\end{figure*}

As illustrated in Figure \ref{fig:overview}, our method consists of four principal components: (1) task difficulty scoring, (2) schema subset generation for prompt seeding, (3) refinement of candidate queries via evolutionary mutation operators (see figure \ref{fig:schemasubsets}), and (4) self-consistency voting for choosing a final solution. The following sections describe each component in detail. The pseudocode for our method can be found in  \ref{alg:algorithm}.

\subsection{Task Difficulty Scoring}

Central to the concept of scaling inference-time compute is the task difficulty scoring component. Given the natural language task $Q$, full database schema $S$, and usage example $H$, we use an LLM to categorize the question as belonging to one of five difficulty levels. Thus, we generate a score $C = LLM_{difficulty}(Q, S, H)$, $1 \leq C \leq 5$. In practice, the upper and lower complexity score ranges can be adjusted to meet specific budgetary requirements. Here, however, we chose the lower threshold as $1$ and the upper threshold as $5$. The LLM is prompted to produce an integer value in the range of the upper and lower thresholds. The resulting score is used to control both the number of parallel refinements performed on candidate queries and the depth of each regiment reasoning path. In this way, inference-time compute resources can be dynamically allocated, with more challenging questions receiving greater amounts of reasoning time than less complex tasks.

\subsection{Schema Subset Generation for Prompt Seeding}

A critical step in the Text-to-SQL process is the schema linking stage, whereby the tables and columns most relevant to the task are selected from the full database schema. Proper schema linking has been shown to improve the accuracy of Text-to-SQL pipelines by providing the candidate SQL query generator with only the most pertinent schema elements, reducing rates of hallucination \cite{sl1}. Techniques for performing schema linking abound, including single schema subset selection \cite{OPENSEARCH}, repeated sampling then merging the results \cite{MCS}, and even forgoing the process entirely and simply providing the query generator LLM with the entire schema \cite{death-schema-linking}. In each of these instances, regardless of the approach, only a single subset $S’ \subseteq S$ is derived and then used across all future query generation calls. This limits the query generator LLM’s exploration capabilities, as its schematic frame of reference is limited to the same list of tables and columns across all calls. Some solutions attempt to overcome this limitation by randomly swapping the ordering of the tables and columns embedded in the prompt \cite{MCS} or by using higher temperatures \cite{ESQL}, but our experiments demonstrate that this only results in minor increases in query variance, and achieves subpar query uniqueness. As a result, we propose generating a pool of distinct schema subsets, each to be used in the seeding of a different candidate query. Unique schema subsets encourage greater exploration by exposing the query generation LLM to a broader swath of table-column combinations, as opposed to a single static subset. To generate the schema subsets, the subset generator $LLM_{schema\_subset}$ is sampled $N$ times to produce a subset $S’_{i}$ of tables and columns that are most relevant to the question $Q$:

\begin{equation}
    S’_{1}, …, S’_{N} \sim LLM_{schema\_subset}(Q, S, H)
\end{equation}

Each unique subset is stored in the seed pool. After sampling, we merge all the subsets into a superset and add it to the pool. To validate our hypothesis that using a pool of schema subsets leads to greater variety in the queries produced as opposed to a single schema subset, we ran query generation experiments across random samples of tasks from the BIRD benchmark. We found that query pools generated from the pool of seeds contained a greater number of unique, diverse queries as opposed to pools with the same number of queries, all generated from the same, single subset seed. The details of this experiment can be found in Section 4.6.

In sum, we create a pool P of schema subsets $S’_{i}$, where each subset $S’_{i}$ is used to generate a candidate query in subsequent components of the pipeline.

\subsection{Candidate Query Generation and Refinement}

For a given schema subset $S’_{i}$, a candidate query $QC$ is produced by prompting the query candidate generator:

\begin{equation}
    QC = LLM_{gen\_query}(Q, S'_{i})
\end{equation}

To evaluate its accuracy, $QC$ is then run on the database via the query execution module $f_{execute}$, which returns values $(E, O)$, where $E$ consists of any errors that occurred when running the query on the database and $O$ is the output produced by the query:

\begin{equation}
    (E, O) = f_{execute}(QC)
\end{equation}

Next, the query critic LLM is tasked with evaluating the candidate query and providing an assessment of how it can be improved. Due to the large number of candidate queries to evaluate within the Text-to-SQL pipeline, any queries which have raised an error are eliminated from the candidate pool, in order to reduce compute costs, thus, only queries that ran successfully are passed to the critic. The critic $LLM_{critic}$ is given the schema subset $S’_{i}$, question $Q$, candidate query $QC$, output $O$, evidence hint $H$, and is prompted to produce a score $s$, $0 \leq s \leq 1$, confidence $k$, $0 \leq k \leq 1$, mutation temperature $t$, $0 \leq t \leq 1$, and an assessment $A$:

\begin{equation}
    (s, k, t, A) = LLM_{critic}(Q, S’_{i}, H, QC, O)
\end{equation}

Below is a description of each parameter and its usage in our pipeline:

\textbf{Score}: The critic is tasked to provide a score that reflects how well the candidate query solves the task by examining the output and determining how closely it matches the intent of the question. The greater the score, the greater the likelihood the candidate query has produced the correct response.

\textbf{Confidence Score}: The critic is asked to produce a confidence score, reflecting any uncertainties that may exist in its prediction. This acts as a weight for the candidate query reward computation (see Section \ref{queryscore}).

\textbf{Mutation Temperature}: The mutation temperature reflects the degree to which the critic believes the query needs to be changed to better solve the task. The greater the temperature, the greater the number of changes needed. We use this score in our reward calculation as a tiebreaker, so that queries with the same or similar scores can be better distinguished by the number of changes still needed to improve them. For example, candidates $QC_{1}$ and $QC_{2}$ may both return the same or similar output, thus granting them the same overall score from the critic, however, $QC_{1}$ may require more changes than $QC_{2}$ by virtue of how its query is written. Thus, their overall reward scores should reflect the amount of future refinement needed for improvement.

\textbf{Assessment}: The critic is prompted to produce a recommendation of what changes should be made to the query so that the correct output will be produced. If no further changes are necessary, the assessment is left blank.

\subsection{Candidate Query Reward}\label{queryscore}

From the critic’s responses, a reward score for the candidate query $QC$ is calculated:

\begin{equation}
    R = s * (1 - t) * k
\end{equation}

Here, the overall query score $s$ is weighted by both the confidence score and mutation temperature. $t \to 1$ as the query requires more changes, thus $(1 - t) \to 1$ as the query increases in accuracy. 

\subsection{Query Mutation}

Given the mutation temperature $t$ and assessment $A$ from the critic, the original candidate $QC$ is then rewritten via $LLM_{mutate}$, which is prompted to produce an updated query candidate $QC’$ that incorporates the changes recommended by the critic:

\begin{equation}
    QC’ = LLM_{mutate}(Q, S’_{i}, QC, H, A, t)    
\end{equation}

We consider a single refinement step to consist of a call to the critic, followed by a subsequent call to the mutator. 

\subsection{Scaling Candidate Query Generation and Refinement}

We use the task difficulty score $C$ as the factor by which candidate query generation and refinement is scaled. Specifically, we perform subset seed pool creation, candidate query generation, and refinement of each candidate query in the pool $C$ times, accumulating a buffer $B$ of $(QC, R, O)$ tuples across each refinement step. We set the maximum number of refinement steps for each candidate query to be $\left\lfloor \frac{C}{2} \right\rfloor + 1$. In this way, more complex tasks are given greater refinement depth, while ensuring that the quality of the critic feedback and query mutations does not degrade across too many steps \cite{DEEPMIND}. 

At each iteration over $C$, we generate a new pool $P$ of schema subsets. Instead of resampling from $LLM_{schema\_subset}$, we use the evolutionary search operators of crossover and mutation to produce new schema subsets. Until the new pool $P’$ has size $|P|$, we choose two subsets from $P$ and merge their contents (crossover) with probability $p$ and choose a single subset with probability $1-p$ and randomly remove a table-column pair from it until the resulting subset has not been observed before.

\subsection{Final Answer Selection}

After performing the search for solution queries over $C$ iterations, we use the maximum sum of rewards to choose the final output query. We first create a mapping $F : \{ O \mid (QC, R, O) \in B \} \to \{ QC \mid (QC, R, O) \in B \}$ which links each observed output $O$ to a corresponding query $QC$. Then, the output with the greatest sum of rewards is selected:

\begin{equation}
    O_{solution} = \arg\max_{O'} \sum_{(QC, R, O) \in B} R [O = O']
\end{equation}

Finally, the query associated with the output $O_{solution}$ is returned:

\begin{equation}
    QC_{solution} = f(O_{solution})
\end{equation}

We experimented with several different self-consistency metrics, including majority voting, but our proposed sum-of-rewards approach performed the best. Please see Section \ref{ablationstudy} for ablation details. 

\section{Experiments} 

\subsection{Datasets}

We evaluate the performance of CA-SQL on the BIRD benchmark, the most complex and comprehensive cross-domain Text-to-SQL dataset. BIRD contains 95 large, anonymized databases across 37 real-world domains. In particular, BIRD Text-to-SQL tasks are frequently noisy and dirty, necessitating advanced reasoning to interpret \cite{Wretblad2024-qs}. We ran the CA-SQL pipeline on each Text-to-SQL task in BIRD's development dataset, which contains 1534 natural-language tasks and their corresponding gold queries for accuracy scoring. 

\begin{table*}[!h]
    \centering
\begin{tabular}{lcccccccl}\toprule
& \multicolumn{2}{c}{\textbf{Simple}} & \multicolumn{2}{c}{\textbf{Moderate}} & \multicolumn{2}{c}{\textbf{Challenging}} & \multicolumn{2}{c}{\textbf{Overall}} \\
\cmidrule(lr){2-3}\cmidrule(lr){4-5}\cmidrule(lr){6-7}\cmidrule(lr){8-9}
  \textbf{Method}         & \textbf{EX} & \textbf{Soft F1} & \textbf{EX} & \textbf{Soft F1} & \textbf{EX} & \textbf{Soft F1} & \textbf{EX} & \textbf{Soft F1}  \\

\midrule
MCS + GPT-4 & 70.4 & - & 53.1 & - & 51.4 & - & 63.4 & -\\
E-SQL + GPT-4o-mini & 67.44 & 68.8 & 56.94 & 58.77 & 40 & 43.04 & 59.81 & 61.59\\
E-SQL + GPT-4o & 73.02 & 73.91 & 64.14 & 66.17 & 48.07 & 51.45 & 66.29 & 67.93\\
MCTS + GPT-4o-mini & 68.56 & - & 57.76 & - & 45.83 & - & 63.15 & -\\
MCTS + GPT-4o & 74.32 & - & 65.17 & - & 51.48 & - & 69.4 & -\\
\midrule
\textbf{Ours + GPT-4o-mini} & 66.93 & \textbf{74.61} & 52.3 & \textbf{60.73} & \textbf{51.72} & \textbf{57.23} & 61.06 & \textbf{68.77}\\
\bottomrule
\end{tabular}
\caption{Performance of CA-SQL on the BIRD development set.}
\label{tab:performance}

\end{table*}

\subsection{Evaluation Metrics}

Performance is gauged via BIRD’s Execution Accuracy (EX) and Soft F1 scores. Execution accuracy is the percentage of queries that produce the same output as their corresponding task’s gold query. Soft F1, however, is a more flexible metric that is designed to account for minor discrepancies between the gold output and the framework’s output. This allows the performance evaluation to account for outputs that largely meet the intent of the original task, but contain small variations in column ordering, value presence, etc. 

\subsection{Models}
In all our experiments and ablations, we used GPT-4o-mini as our base model. Our primary purpose in using GPT-4o-mini was to test the efficacy of a small, general purpose LLM when combined with our in-context learning techniques and inference-time scaling approach, as opposed to using larger models. 

\subsection{Hyper-Parameters}

When running CA-SQL on a task, we set the maximum number of calls $N$ to $LLM_{schema\_subsets}$ to $20$. The crossover probability $p$ was set to $50\%$. The temperature of all LLMs can be found in Table \ref{tab:hyperparams}.

\begin{table}[!h]
    \centering
    \begin{tabular}{lr}
        \toprule
        Model  & Temperature \\
        \midrule
        $LLM_{schema\_subset}$     & 1.0  \\
        $LLM_{gen\_query}$  & 1.0  \\
        $LLM_{critic}$ & 0.2  \\
        $LLM_{mutate}$ & 1.0  \\
        $LLM_{difficulty}$ & 1.0 \\
        \bottomrule
    \end{tabular}
    \caption{CA-SQL LLM model temperatures}
    \label{tab:hyperparams}
\end{table}

\subsection{Results}

In Table \ref{tab:performance}, we list comparisons between the performance of CA-SQL and other solutions that utilize in-context learning (ICL). For the sake of thoroughness, we include both EX and Soft F1 scores, if available. Our solution outperforms all other methodologies on the “challenging” difficulty level of BIRD, scoring a \textbf{51.72\%} execution accuracy (EX). This score is particularly notable, as it was achieved using GPT-4o-mini as the base, much smaller and cheaper than the GPT-4o and GPT-4 models utilized by other solutions. Our result on the challenging tasks is a \textbf{0.24\%} increase over the current SOTA using GPT-4o and a \textbf{5.89\%} increase over the current SOTA using GPT-4o-mini. This result demonstrates that proper exploration is vital to improving the performance of LLMs on challenging problems, especially when using smaller models. This further shows that smaller models can outperform larger models that apply SOTA inference-time tree search methodologies, such as MCTS. CA-SQL’s Soft F1 scores across the three levels of difficulty outperform all reported Soft F1 scores for ICL solutions using GPT-4o-mini. Also, our overall Soft F1 score is \textbf{0.83\%} better than E-SQL + GPT-4o, representing commensurate performance with solutions using larger models. 

It is noteworthy, however, that CA-SQL’s improved score on challenging problems is not observed across the simple and moderate task tiers, as we perform slightly worse than current SOTA techniques on those problem categories. This is because these other solutions prioritize depth-based candidate query refinement, as opposed to devoting more inference-time compute resources to exploration. Even MCTS, designed to balance the tradeoff between exploration and exploration, still falls short, as the gap in performance between our solution on challenging problems and MCTS’ performance illustrates. We perform a detailed analysis of this phenomenon in Section \ref{limmcts} of Appendix~\ref{sec:appendix}.

\section{Discussion}

The results from the ablation study (see \ref{ablationstudy} in Appendix~\ref{sec:appendix}) are in keeping with the observations from prior studies \cite{Liu2025-ol,Parashar2025-nl,DEEPMIND,Zhang2025-ad,Chen2025-fk} on inference-time scaling techniques. Our exploration-based methodology improves our pipeline’s performance on moderate and challenging text-to-SQL problems, but does not demonstrate commensurate performance on simpler problems, where increased depth-based refinement is required. Optimal techniques for depth-wise candidate query refinement are out of the scope of this work, as we aim to show that scaling the number of candidate queries leads to improved performance on difficult tasks. In addition, sum-of-rewards voting outperforms all other methods tested. 

\section{Conclusion}

In this paper, we propose CA-SQL, a text-to-SQL pipeline that scales solution exploration to match the estimated difficulty of a given task. We also demonstrate that increased coverage of the candidate query space can be achieved by utilizing our novel schema subset generator and evolutionary search-inspired operators. Lastly, we use a custom sum-of-rewards voting strategy to choose a final query to return. Our experiments and ablations demonstrate that CA-SQL outperforms the current state of the art in-context learning solutions on the “challenging” set of BIRD benchmark tasks and achieves competitive accuracy on the rest, with our difficulty-based scaling technique proving to be a vital contributor to that accuracy increase.

\section*{Limitations}

Due to time and budgetary constraints, we were only able to test CA-SQL on GPT-4o-mini, as opposed to other larger, proprietary models, such as GPT-4o. In future work, we plan to test our framework on a wider array of models to better generalize our finding that increased search space exploration improves pipeline performance on challenging Text-to-SQL problems. Lastly, further experiments are needed to gauge the variance of a given Text-to-SQL problem's difficulty score. While we sample a difficulty score from $LLM_{difficulty}$ only once, it is possible that task difficulty scores may fluctuate considerably across a vareity of samples from the same model and as such, may require special voting methods for aggregation and selection.

\bibliography{custom}

\appendix

\section{Appendix}
\label{sec:appendix}
\subsection{Ablation Study}\label{ablationstudy}

We conducted an ablation study to gauge the effectiveness of our solution. We ran CA-SQL on BIRD first without scaling to meet each task’s estimated difficulty, and then with our proposed scaling mechanism. Furthermore, we compared the performance of different voting mechanisms against our proposed sum-of-rewards approach. The ablation results (see Table \ref{tab:ablations}) demonstrate that scaling improves performance on moderate and challenging problems, with the latter problems possessing the largest performance increase of 1.4\% EX. For voting methods, our sum-of-rewards voting strategy led to significantly better performance than majority voting (+ 3.4\% EX overall), highest average reward (+ 7.77\% EX overall), and highest scoring (+ 1.6\% EX overall).

\begin{table*}[!h]
    \centering
\begin{tabular}{lcccccccc}\toprule
& \multicolumn{2}{c}{\textbf{Simple}} & \multicolumn{2}{c}{\textbf{Moderate}} & \multicolumn{2}{c}{\textbf{Challenging}} & \multicolumn{2}{c}{\textbf{Overall}} \\
\cmidrule(lr){2-3}\cmidrule(lr){4-5}\cmidrule(lr){6-7}\cmidrule(lr){8-9}
  \textbf{\textbf{Voting Method (+ Scaling)}}         & \textbf{EX} & \textbf{Soft F1} & \textbf{EX} & \textbf{Soft F1} & \textbf{EX} & \textbf{Soft F1} & \textbf{EX} & \textbf{Soft F1}  \\

\midrule
Highest average reward (w/o scaling) & 62.15 & 68.79 & 45.3 & 54.98 & 44.14 & 50.96 & 55.35 & 62.93\\
Majority voting (w/o scaling) & 64.71 & 73.1 & 47.05 & 55.74 & 45.52 & 52.23 & 57.55 & 65.88\\
Highest reward (w/o scaling) & 64.59 & 71.62 & 49.89 & 58.47 & 47.59 & 54.41 & 58.54 & 66.02\\
Sum of rewards (w/o scaling) & \textbf{67.48} & \textbf{74.79} & 50.55 & 58.99 & 50.34 & 56.8 & 60.74 & 68.31\\
\midrule
Highest average reward (scaling) & 60.38 & 67.26 & 42.01 & 52.85 & 44.14 & 50.76 & 53.29 & 61.34\\
Majority voting (scaling) & 63.37 & 71.49 & 49.45 & 57.87 & 47.59 & 54.92 & 57.67 & 65.8\\
Highest reward (scaling) & 65.93 & 72.84 & 50.11 & 58.82 & 48.28 & 55.46 & 59.47 & 66.96\\
Sum of rewards (scaling) & 66.93 & 74.61 & \textbf{52.3} & \textbf{60.73} & \textbf{51.72} & \textbf{57.23} & \textbf{61.06} & \textbf{68.77}\\
\midrule
Sum of rewards (w/o scaling) & \textbf{67.48} & \textbf{74.79} & 50.55 & 58.99 & 50.34 & 56.8 & 60.74 & 68.31\\
\textbf{Sum of rewards (scaling)} & 66.93 & 74.61 & \textbf{52.3} & \textbf{60.73} & \textbf{51.72} & \textbf{57.23} & \textbf{61.06} & \textbf{68.77}\\
\bottomrule
\end{tabular}
\caption{Ablation study of self-consistency voting methods in conjunction with search breadth scaling control.}
\label{tab:ablations}

\end{table*}

\subsection{Limitations of MCTS and Exploration Analysis}\label{limmcts}

We believe the underperformance of MCTS on challenging problems can be attributed to deficient candidate query sampling techniques. The fundamental exploratory aspect of MCTS is node expansion, where for a given candidate solution $S$, child solutions $C_{1}, C_{2},..., C_{n}$ are sampled from an LLM that has been given the contextual details of $S$. This limits the diversity of the samples $C_{1}, C_{2},..., C_{n}$, as they all originate from the same prompt \cite{LIGHT}. To demonstrate this lack of diversity in a Text-to-SQL context, we performed a comparison of three different candidate query generation techniques: (1) generation from a pool of schema subsets (our approach), (2) generation from a single schema subset with random ordering of tables and columns in the prompt (the conventional approach), and (3) generation from a single schema subset with static ordering of tables and columns (default control). We first randomly sampled 100 tasks from each of the three BIRD difficulty levels. For each selected task, using our sampling parameter $N=20$, we created a pool $P$ of $N$ unique schema subsets via our previously described method, and then applied $LLM_{gen\_query}$ to each. Then, we merged the subsets into one schema superset $S_{super}$ that would serve as the seed for methods (2) and (3), ensuring that those methods would receive a schema subset that provided a large degree of coverage across potentially relevant tables and columns. Then, we sampled $C’_{1}, …, C’_{N} \sim LLM_{gen\_query}(Random\_Order(S_{super}))$ and $C^{\prime\prime}_{1}, …, C^{\prime\prime}_{N} \sim LLM_{gen\_query}(S_{super})$, and stored $C’$ and $C^{\prime\prime}$ in pools $P’$ and $P^{\prime\prime}$, respectively. Lastly, we computed $\frac{|set(p)|}{|p|}$, for all $p$ in ${P, P’, P^{\prime\prime}}$. We repeated the experiment for $LLM_{gen\_query}$ temperatures of $0.0$, $0.3$, $0.7$, and $1.0$. Our results are contained in Table \ref{tab:sampling}. Our subset seeding approach yields a greater percentage of unique queries in the resulting candidate pool than other two solutions, demonstrating that significant variance in prompt composition is necessary to elicit exploratory behavior from an LLM. As current MCTS approaches in Text-to-SQL utilize sampling method (2), it is not surprising that its search tree’s children may lack the diversity of candidate queries needed to adequately cover the solution search space. 

\begin{table*}[!h]
    \centering
\begin{tabular}{lcccc}\toprule
& \multicolumn{4}{c}{\textbf{Percentage of Unique Queries After Sampling}}\\
\cmidrule(lr){2-5}  \textbf{Sampling Method}         & $Temp = 0.0$ & $Temp = 0.3$ & $Temp = 0.7$ & $Temp = 1.0$\\

\midrule
Single seed, fixed ordering & 31.03 & 41.67 & 61.02 & 68.07\\
Single seed, random ordering & 58.46 & 62.97 & 70.17 & 76.66\\
\midrule
\textbf{Pool of seeds (ours)} & \textbf{84.44} & \textbf{86.06} & \textbf{86.97} & \textbf{90.89}\\
\bottomrule
\end{tabular}
\caption{Percentage of unique queries in candidate pool for various query sampling methods at varying temperatures, including our proposed pool of seeds method.}
\label{tab:sampling}

\end{table*}

\begin{algorithm}[!h]
    \caption{The CA-SQL algorithm}
    \label{alg:algorithm}
    \textbf{Input}: question $Q$, schema $S$, usage hint $H$\\
    \textbf{Parameters}: sampling calls $N=20$, crossover probability $p=0.5$\\
    \textbf{Output}: SQL query $QC_{solution}$
    \begin{algorithmic}[1] 
        \STATE $C = LLM_{difficulty}(Q, S, H)$
        \STATE $S’_{1}, …, S’_{N} \sim LLM_{schema\_subset}(Q, S, H)$
        \STATE $P = \{S’_{1}, …, S’_{N}\}$
        \STATE $P = P \cup \bigcup\limits_{S_{i}\in P}^{} S_{i}$
        \STATE $B = \{\}$
        \WHILE{$C \ge 0$}

        \STATE $queue = \{(S'_{i}, 0) \mid S'_{i} \in P\}$

        \WHILE{$|queue| > 0$}
        \STATE $(S'_{i}, depth) = queue.popleft()$
        \STATE $QC = LLM_{gen\_query}(Q, S'_{i})$
        \STATE $(E, O) = f_{execute}(QC)$
        \STATE $(s, k, t, A) = LLM_{critic}(Q, S’_{i}, H, QC, O)$
        \STATE $R = s * (1 - t) * k$
        \STATE $QC’ = LLM_{mutate}(Q, S’_{i}, QC, H, A, t)$
        \STATE $B = B \cup \{(QC, R, O)\}$
        
        \IF {$depth < \left\lfloor \frac{C}{2} \right\rfloor + 1$}
        \STATE $queue = queue \cup \{(QC', depth + 1)\}$
        \ENDIF
        
        \ENDWHILE
        
        \STATE $ C = C - 1$
        
        \STATE $P' = \{\}$
        \WHILE{$|P'| < |P|$}
        \STATE $S', S^{\prime\prime} \sim P$
        \IF {$\mathcal{U}(0, 1) \le p$}
        \STATE $P' = P\cup(S' \cup S^{\prime\prime})$
        \ELSE
        \STATE $S' \sim P$
        
        \WHILE{$S' \in P$}
        \STATE $(T, C) \sim S'$
        \STATE $S' = S' \setminus {(T, C)}$
        \ENDWHILE
        \STATE $P' = P' \cup \{S'\}$
        \ENDIF
        \ENDWHILE
        \STATE $P=P'$
        \ENDWHILE
        \STATE $F: \{ O \mid (QC, R, O) \in B \} \to \{ QC \mid (QC, R, O) \in B \}$
        \STATE $O_{solution} = \arg\max_{O'} \sum_{(QC, R, O) \in B} R [O = O']$
        \STATE $QC_{solution} = f(O_{solution})$
        \STATE \textbf{return} $QC_{solution}$
    \end{algorithmic}
\end{algorithm}

\end{document}